\documentclass[10pt,twocolumn,letterpaper]{article}

\usepackage[pagenumbers]{iccv} 

\definecolor{iccvblue}{rgb}{0.21,0.49,0.74}
\usepackage[pagebackref,breaklinks,colorlinks,allcolors=iccvblue]{hyperref}
\usepackage{tcolorbox}
\newtcolorbox{promptbox}{
  colback=gray!10,      
  colframe=gray!50,     
  rounded corners,      
  boxrule=0.5pt,        
  width=\linewidth,     
}
\newtcolorbox{qabox}{
  colback=gray!10,      
  colframe=gray!50,     
  rounded corners,      
  boxrule=0.5pt,        
  width=\linewidth,     
}
\usepackage{algorithm}
\usepackage{algorithmic}
\usepackage{pifont}

\title{SpatialPrompting: Keyframe-driven Zero-Shot Spatial Reasoning with Off-the-Shelf Multimodal Large Language Models}

\author{Shun Taguchi \ Hideki Deguchi \ Takumi Hamazaki \ Hiroyuki Sakai \\
Toyota Central R\&D Labs., Inc.\\
}

\begin{document}
\maketitle

\begin{abstract}
This study introduces SpatialPrompting, a novel framework that harnesses the emergent reasoning capabilities of off-the-shelf multimodal large language models to achieve zero-shot spatial reasoning in three-dimensional (3D) environments. Unlike existing methods that rely on expensive 3D-specific fine-tuning with specialized 3D inputs such as point clouds or voxel-based features, SpatialPrompting employs a keyframe-driven prompt generation strategy. This framework uses metrics such as vision–language similarity, Mahalanobis distance, field of view, and image sharpness to select a diverse and informative set of keyframes from image sequences and then integrates them with corresponding camera pose data to effectively abstract spatial relationships and infer complex 3D structures. The proposed framework not only establishes a new paradigm for flexible spatial reasoning that utilizes intuitive visual and positional cues but also achieves state-of-the-art zero-shot performance on benchmark datasets, such as ScanQA and SQA3D, across several metrics. The proposed method effectively eliminates the need for specialized 3D inputs and fine-tuning, offering a simpler and more scalable alternative to conventional approaches.
\end{abstract}

\section{Introduction}
\label{sec:introduction}
\begin{figure}
    \centering
    \includegraphics[width=1.0\linewidth]{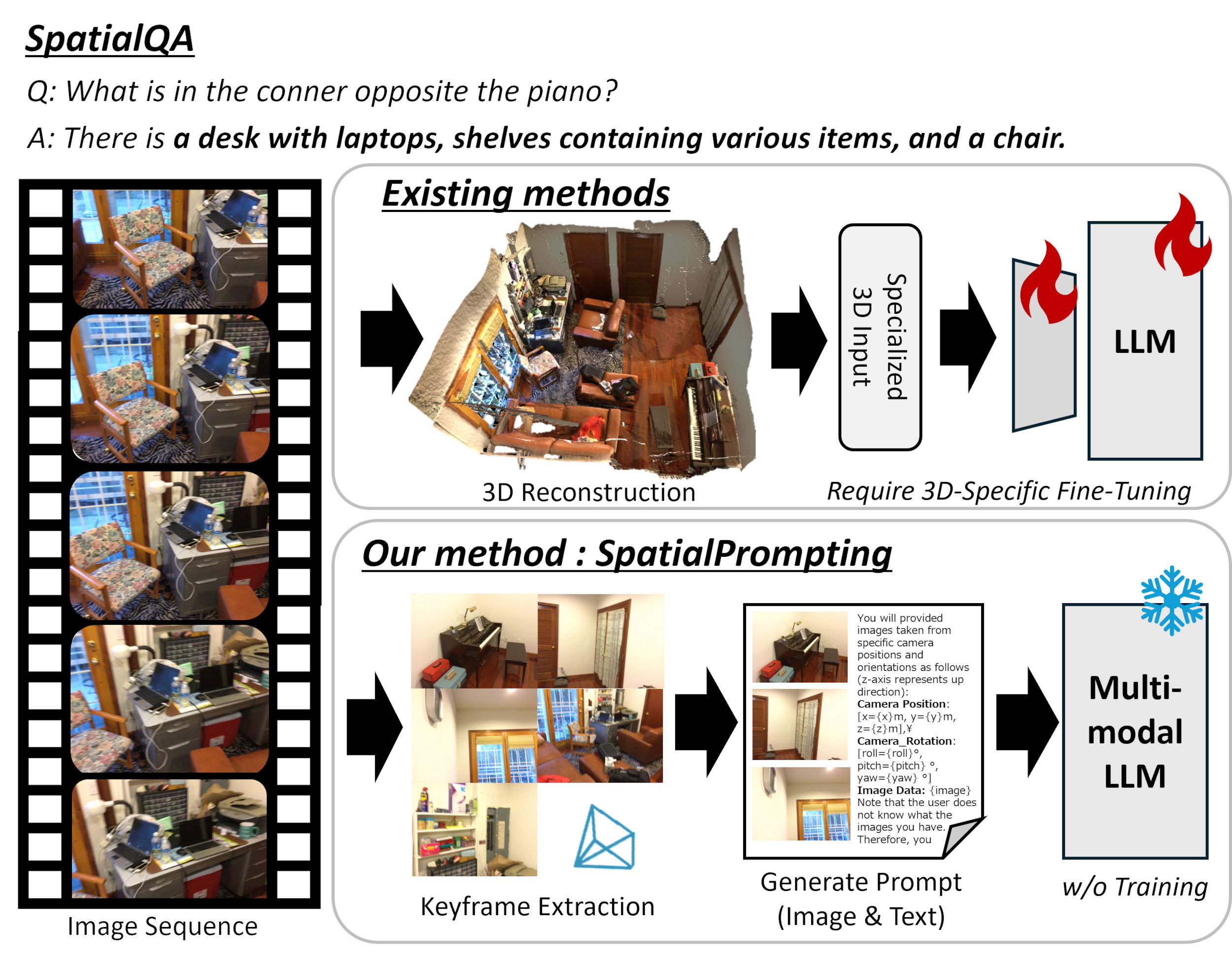}
    \caption{{\bf SpatialPrompting} is a framework that employs keyframe-driven prompt generation for zero-shot spatial reasoning with multimodal LLMs. The proposed approach enables accurate spatial reasoning without additional 3D-specific training.}
    \label{fig:pull_figure}
\end{figure}

Spatial question answering (SpatialQA) in a three-dimensional (3D) environment is essential in computer vision because it enables machines to understand, reason about, and interact with visual information. For instance, home assistant systems rely on accurate spatial understanding to provide users with relevant information about room layouts and object locations---such as helping them locate misplaced items or offering interior organization guidance. Similarly, service robots performing indoor household tasks depend on precise spatial reasoning to operate effectively. Furthermore, robust spatial reasoning is indispensable for interactive applications such as augmented reality, where seamlessly integrating digital content with physical space enhances user experiences across various domains.

Several attempts have already been made to achieve SpatialQA. Previous methods~\cite{azuma2022scanqa, ma2022sqa3d, parelli2023clip} focused on task-specific models that utilized 3D features such as point cloud representations. Recently, methods that integrate large language models (LLMs) and learning encoders for 3D feature representations and fine-tuning the LLMs have been proposed. For instance, scene-LLM~\cite{fu2024scene} adopts voxel-based 3D features, chat-scene~\cite{huang2024chat} introduces object-centric representations, and 3D-graph LLM~\cite{zemskova20243dgraphllm} introduces graph structures to capture the relationships between objects. These advances, achieved through the integration of multiple modalities and sophisticated learning strategies, have led to significant improvements in SpatialQA performance, although they often occur at the expense of expensive 3D-specific fine-tuning.

Unlike conventional methods that rely on expensive 3D-specific fine-tuning, we propose a training-free approach that enables SpatialQA by leveraging the capabilities of multimodal LLMs. Multimodal LLMs, such as GPT-4o~\cite{openai2024gpt4o}, are pre-trained on vast amounts of data, enabling them to learn abstract concepts and demonstrate few-shot learning as well as emergent capabilities~\cite{wei2022emergent, brown2020language}. Recent studies have revealed that LLMs can achieve video reasoning through prompt engineering~\cite{wang2024videotree, liao2024videoinsta}, suggesting that these advanced reasoning skills can be extended to spatial relationship abstraction. We hypothesize that through appropriate in-context learning from a few images, effective spatial reasoning can be performed using only images captured from different perspectives along with their corresponding camera pose data. This eliminates the need for 3D-specific fine-tuning.

Therefore, we introduce SpatialPrompting, a novel framework that leverages the emergent reasoning capabilities of off-the-shelf multimodal LLMs.
Rather than relying on expensive 3D-specific fine-tuning, the proposed approach utilizes a representative set of keyframes extracted from image sequences---each accompanied by corresponding camera pose data---to perform effective zero-shot spatial reasoning (\cref{fig:pull_figure}). 
By employing a keyframe-driven prompt generation strategy, the proposed method prunes redundant views using metrics such as vision–language similarity and Mahalanobis distance while prioritizing images with a broader field of view (FOV) and higher sharpness.
This seamless integration of visual and positional cues empowers multimodal LLMs to accurately infer complex spatial structures without the overhead of 3D-specific fine-tuning, enabling scalable and cost-effective applications in real-world 3D environments.

In summary, our contributions are threefold: 
\begin{itemize} 
    \item {\bf Spatial reasoning based on multimodal LLM capabilities:} We introduce a new paradigm to SpatialQA by leveraging intuitive visual and positional cues, thereby eliminating the dependency on expensive 3D-specific fine-tuning.
    \item {\bf Keyframe-driven prompt generation:} We develop an efficient keyframe selection mechanism that minimizes redundancy while maximizing informational content, thereby facilitating effective zero-shot spatial reasoning with multimodal LLMs. 
    \item {\bf State-of-the-art (SOTA) zero-shot performance:} We demonstrate that, even without additional fine-tuning, the proposed approach achieves competitive performance on ScanQA~\cite{azuma2022scanqa} and SQA3D~\cite{ma2022sqa3d}, enabling scalable and cost-effective applications in real-world 3D environments. 
\end{itemize} 
This study establishes a promising new direction for 3D spatial reasoning, challenging the conventional wisdom that high-dimensional, specialized inputs are necessary for effective spatial understanding, and marking a significant step toward more natural and intuitive machine perception.

\section{Related Work}
\label{sec:related_work}

\noindent{\bf Spatial-language understanding}
In recent years, language information has been actively utilized in spatial understanding to accurately capture user intentions. Representative tasks include 3D visual grounding~\cite{chen2020scanrefer, achlioptas2020referit3d, huang2022multi, zhang2023multi3drefer, chen2022language, wang20233drp, zhao20213dvg, wang2023distilling, unal2024four}, which identifies targets within a 3D scene based on text queries; 3D dense captioning~\cite{chen2021scan2cap, yuan2022x, jiao2022more, chen2023end, chen2024vote2cap}, which localizes objects in the scene and provides detailed descriptions; and 3D visual question answering (3D-VQA)~\cite{azuma2022scanqa, ma2022sqa3d, parelli2023clip}, which responds to queries about the entire scene. Previous studies primarily focused on task-specific methods, but with the advent of pre-training approaches such as 3D-VisTA~\cite{zhu20233d} and 3D-VLP~\cite{jin2023context}, the attention has shifted towards leveraging commonalities across tasks. However, the reliance on task-specific heads remains a challenge for broader user interaction applications.
With the evolution of LLMs, methods that integrate 3D data for spatial understanding have emerged. Approaches such as 3D-LLM~\cite{hong20233d} and Chat-3D~\cite{wang2023chat} employ techniques such as positional embedding and pre-alignment to enhance the accuracy of 3D scene understanding. Moreover, methods such as scene-LLM~\cite{fu2024scene}, which employs voxel-based 3D features, chat-scene~\cite{huang2024chat}, which adopts an object-centric representation, and 3D-graph LLM~\cite{zemskova20243dgraphllm}, which utilizes graph structures to capture relationships among objects, have each made distinct advances. Although the fusion of multiple modalities and advanced learning strategies has significantly improved the performance of SpatialQA, it has underscored the issue of expensive, 3D-specific fine-tuning requirements.

\noindent{\bf In-context learning and LLM capabilities}
Recent advances in in-context learning have significantly enhanced the ability to adapt to tasks using few-shot examples. As demonstrated by Brown et al.~\cite{brown2020language}, LLMs can extract task-specific patterns and rules from natural language instructions and a few demonstrations, without the need for dedicated fine-tuning. 
This capability has been further validated in some studies~\cite{mao2023large, kojima2022large}, which reveal that LLMs maintain high performance even in tasks that require extended context or complex reasoning.
These few-shot prompting strategies have also been applied to video question answering. Although video understanding by LLMs has been explored in several studies~\cite{xu2023retrieval, maaz2023video, jin2024chat, yu2023self, lin2023video, zhang2023video, huang2024vtimellm, wang2024vamos}, a significant challenge remains---the learning cost increases substantially as the video length increases.
In this regard, studies have focused on video understanding without requiring additional learning by combining large-scale caption generation, iterative keyframe selection, and information retrieval~\cite{wang2022language, zhang2023simple, wang2024videoagent, choudhury2023zero, min2024morevqa, mangalam2023egoschema, shang2024traveler, wang2024videotree, liao2024videoinsta}.
However, the application of multimodal LLMs' spatial reasoning capabilities by multimodal prompt to achieve SpatialQA from videos has not yet been thoroughly investigated. Therefore, this study aims to directly enhance the spatial reasoning ability of multimodal LLMs by extracting keyframes from videos along with their corresponding camera poses.

\section{Method}
\label{sec:method}

\subsection{Overview}
\label{sec:overview}
\begin{figure*}
    \centering
    \includegraphics[width=0.95\linewidth]{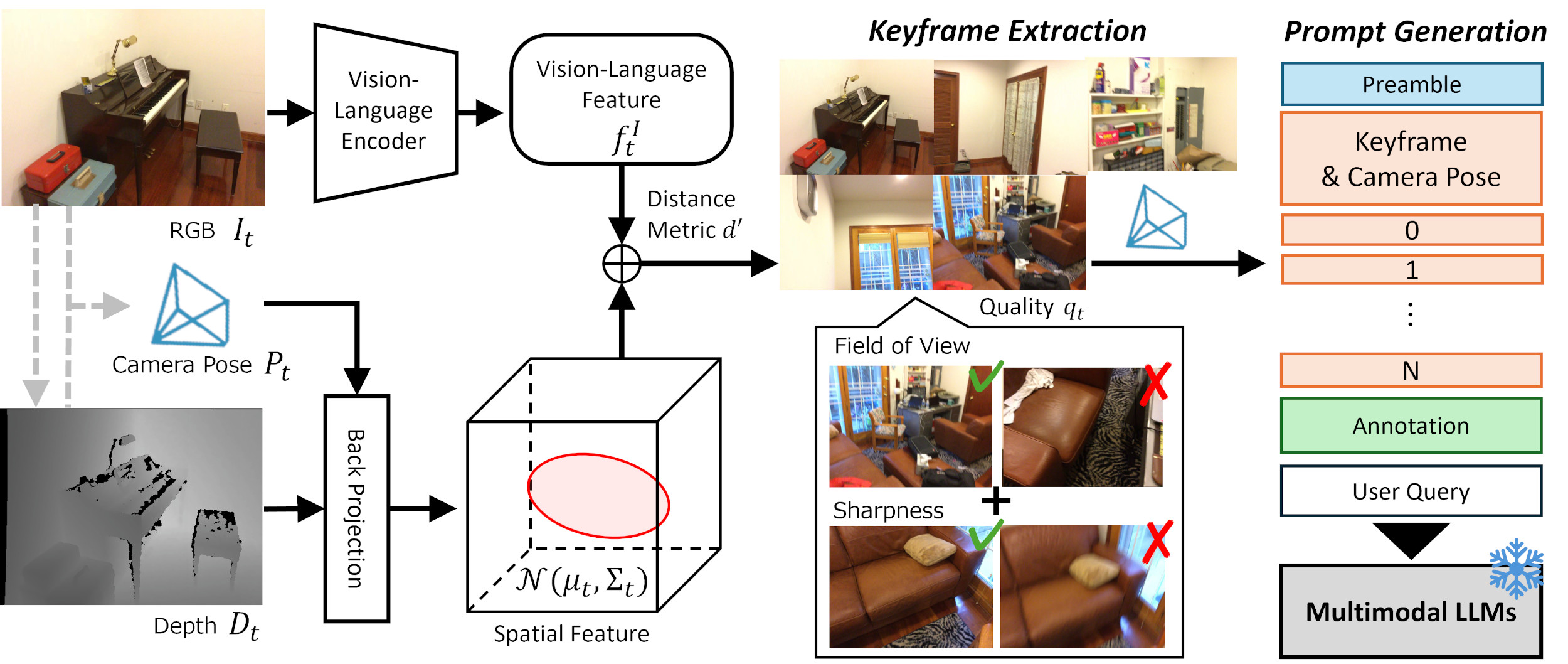}
    \caption{{\bf Overview of SpatialPrompting.} In keyframe extraction, both spatial and semantic features are used to select keyframes. In prompt generation, these keyframes and camera poses are combined with a preamble, annotation, and user query to form prompts for multimodal LLMs, enabling SpatialQA.}
    \label{fig:architecture}
\end{figure*}

The proposed approach is designed to facilitate SpatialQA by leveraging key images extracted from video data. 
The keyframes, along with their associated camera poses, are fed as prompts into multimodal LLMs to impart spatial context. The overall framework comprises two major components: \textit{keyframe extraction} and \textit{prompt generation}.
An overview of the SpatialPrompting framework is shown in \cref{fig:architecture}.

\subsection{Keyframe Extraction from Video}
\label{sec:key_image_extraction}

This section describes the proposed approach for extracting keyframes from video sequences, enabling multimodal LLMs to gain an effective understanding of the environment.

\noindent \textbf{Input data acquisition:}  
When using an RGB-D camera, both RGB images \(I_t\) and depth maps \(D_t\) are directly acquired at each time step \(t\). In scenarios where only RGB data are available, we employ monocular depth estimation techniques (e.g., depth anything~\cite{yang2024depth}) to infer the depth map \(D_t\). With the depth information, the corresponding 3D camera pose \(P_t \in \mathbb{R}^{4 \times 4}\) is estimated using RGB-D SLAM methods such as DROID-SLAM~\cite{teed2021droid} or BundleFusion~\cite{dai2017bundlefusion}.

\noindent \textbf{Feature extraction from video data:}  
The proposed keyframe extraction procedure leverages both semantic and spatial cues by computing feature representations for each frame and evaluating their mutual distances.

\begin{itemize}
    \item \textbf{Vision-language feature extraction:}  
    Vision-language features for each frame are computed using a vision-language model (VLM), such as CLIP~\cite{radford2021learning}.
    \[
    f_t^I = \mathcal{F}_{I}(I_t)
    \]
    Here, \(f_t^I\) denotes the feature representation for frame \(I_t\), and \(\mathcal{F}_{I}\) is the image encoder component of the VLM.

    \item \textbf{Spatial feature extraction:}  
    A 3D point cloud \(Q_t\) is constructed from the depth map \(D_t\) to capture spatial attributes. The back-projection of pixel coordinates \(p = [u, v]^T\) is performed using the intrinsic camera matrix \(K\):
    \[
    Q_t(p) = D_t(p) \, K^{-1} \begin{bmatrix} p \\ 1 \end{bmatrix}.
    \]
    The resulting local point cloud is thereafter transformed into world coordinates via the camera pose \(P_t\).
    \[
    \begin{bmatrix} Q_t^{g}(p) \\ 1 \end{bmatrix} = P_t \begin{bmatrix} Q_t(p) \\ 1 \end{bmatrix},
    \]
where \(Q_t^{g}(p)\) represents the global coordinate of the corresponding point. We further summarize the spatial distribution of \(Q_t^{g}\) by computing its mean \(\mu_t\) and covariance matrix \(\Sigma_t\).
\end{itemize}

\noindent \textbf{Keyframe extraction procedure:}  
Keyframe selection is performed by evaluating a combined distance metric that fuses spatial and semantic similarities, followed by a prioritization step.

\begin{itemize}
    \item \textbf{Mahalanobis distance:}  
    The spatial dissimilarity between two frames \(i\) and \(j\) is measured using the Mahalanobis distance:
    \[
    d(i, j) = (\mu_i - \mu_j)^T \left(\frac{\Sigma_i + \Sigma_j}{2}\right)^{-1} (\mu_i - \mu_j).
    \]
    The Mahalanobis distance calculates the distance between the point clouds of two images, and thus it can be used as an indicator of the extent of overlap of their FOVs.

    \item \textbf{Vision-language similarity:}  
    Semantic similarity is assessed using cosine similarity between the vision-language features:
    \[
    S(i, j) = \frac{f_i^I \cdot f_j^I}{\lVert f_i^I\rVert \, \lVert f_j^I\rVert}.
    \]
    Vision-language similarity is generally used to measure the similarity between images and language, but we utilize the cosine similarity between features computed from images to compare their linguistic meanings.
\end{itemize}

We adopt the Mahalanobis distance because it accurately captures the spatial distribution differences of the 3D point clouds, reflecting variations in viewpoint and scene coverage. In contrast, the vision-language similarity provides semantic insights into the scene content. Combining these two metrics allows us to effectively prune redundant frames by ensuring that both geometric and semantic redundancies are minimized.
The overall distance metric is defined as:
\[
d'(i, j) = d(i, j) + \alpha \bigl(1 - S(i, j)\bigr),
\]
where \(\alpha\) is a weighting parameter that balances the influence of spatial and semantic information. Frames with redundant information (i.e., small \(d'(i, j)\)) are pruned to retain a set of representative keyframes.

Among frame pairs with close distances $d'$, the decision on which one to remove is determined based on the following quality score \(q_t\):
\[
q_t = \det\left|\Sigma_t\right| + \beta \, \operatorname{Var}(\nabla^2 I_t),
\]
where \(\det\left|\Sigma_t\right|\) quantifies the spread of the point cloud, \(\operatorname{Var}(\nabla^2 I_t)\) measures the variance of the Laplacian (reflecting image sharpness), and \(\beta\) is a balancing coefficient. This quality score favors images with a wider FOV and higher clarity. Moreover, this allows in selecting keyframes that maximize spatial coverage while avoiding low-quality images that contain blur.

The keyframe extraction algorithm is summarized in \cref{alg:keyframe_extraction}.
For computational efficiency, both \(d'(i, j)\) for all frame pairs and the quality scores \(q_t\) for all frames are precomputed.

\begin{algorithm}
\caption{Keyframe extraction algorithm}
\label{alg:keyframe_extraction}
\begin{algorithmic}[1]
    \STATE \textbf{Input:} Set of frames \(\{I_1, \dots, I_N\}\), maximum number of keyframes \(N_{\text{max}}\)
    \STATE \textbf{Output:} Selected keyframes \(\{I_k\}\)
    \WHILE{\(|I| > N_{\text{max}}\)}
        \STATE Identify the frame pair \((i, j)\) with the smallest \(d'(i, j)\)
        \IF{\(q_i > q_j\)}
            \STATE Remove frame \(j\)
        \ELSE
            \STATE Remove frame \(i\)
        \ENDIF
    \ENDWHILE
    \STATE \textbf{Return:} Remaining keyframes \(\{I_k\}\)
\end{algorithmic}
\end{algorithm}

\subsection{Prompt Generation for Multimodal LLMs}
\label{sec:prompt_generation}

We generate prompts for multimodal LLMs by leveraging key images together with their corresponding camera positions and orientations obtained through the proposed process. Each prompt comprises four components: a \emph{preamble}, a \emph{keyframe with camera pose}, an \emph{annotation}, and a \emph{user query}.

\noindent{\bf Preamble} \ The preamble introduces the scenario by informing the model that it will receive images captured from specific camera poses within an indoor environment. For instance, the preamble is constructed as follows:
\begin{promptbox}
You will be provided with images captured from specific camera positions and orientations as follows:
\end{promptbox}
\noindent Depending on the application, the role of an LLM agent can be assigned. (e.g., “You are an excellent home assistant system.")

\noindent{\bf Keyframe with camera pose} \ Next, we provide the keyframe images and the corresponding camera's spatial information in the following format:
\begin{promptbox}
{\bf Camera position}: [x=\{x\}m, y=\{y\}m, z=\{z\}m] \\
{\bf Camera rotation}: [x=\{roll\}°, y=\{pitch\}°, z=\{yaw\}°] \\
{\bf Image data}: \{image\}
\end{promptbox}
\noindent This template is a formatted string. Therefore, the placeholders enclosed in braces (\{ \}) are replaced with actual variable values. Our experimentation indicates that explicitly including units for each value and representing angles in Euler format (rather than using quaternions or rotation matrices) produces inputs that are easier for LLMs to interpret. 
Quaternions require four values and are less intuitive for direct textual reasoning, whereas rotation matrices contain redundant information with nine elements. In contrast, Euler angles provide a compact and human-readable representation. Furthermore, as our dataset comprises static scenes without dynamic motion, gimbal lock is not a significant concern in our setting. The image data are formatted according to the specific requirements of the LLM in use.

\noindent{\bf Annotation} \ The annotation is introduced to control the  output of the model in an appropriate manner. In general, the following annotation is introduced to discourage the model from generating responses that directly reference images (e.g., “It is on the right side of the third image”).
\begin{promptbox}
Note that the user does not know the images that you have. Therefore, you should answer the question as concisely as possible without directly referring to the image with words such as “image" or “photo."
\end{promptbox}
\noindent When benchmark datasets containing ground truth are employed, the annotation is utilized as in-context learning to acquaint the model with the inherent characteristics of the dataset.

\noindent{\bf User Query} \ Finally, the prompt concludes with the user query, which may pose any question about the target space. Examples include queries such as ``How many sofas are there in this room?'' or ``What is on top of the sideboard?'' By placing the user query at the end of the prompt, we can respond to the free conversation that follows.

By integrating these structured components into the prompt and supplying them to multimodal LLMs, we effectively enable SpatialQA.

\section{Experiments} 
\label{sec:experiments}
In this section, we evaluate the effectiveness of the proposed SpatialPrompting framework for SpatialQA. The proposed method leverages the inherent spatial reasoning capabilities of multimodal LLMs without relying on specialized 3D input and 3D-specific fine-tuning.

\subsection{Quantitative Results} 
\label{sec:quantitative}

\subsection{Datasets and Evaluation Metrics} 
\label{sec:datasets}
We validate the effectiveness of SpatialPrompting on two challenging benchmarks for 3D-VQA: ScanQA~\cite{azuma2022scanqa} and SQA3D~\cite{ma2022sqa3d}.

\noindent{\bf ScanQA.} \ The ScanQA validation dataset~\cite{azuma2022scanqa} features complex 3D scenes paired with natural language questions. Performance is measured using multiple metrics including exact match at top-1 (EM@1), ROUGE-L, METEOR, CIDEr, and SPICE.

\noindent{\bf SQA3D.} \ The SQA3D dataset~\cite{ma2022sqa3d} features situated questions in complex 3D scenes, such as “I am printing files with the armchair on my left. Which direction do I have to walk to sit down on the couch?”. We categorize the questions into types such as “What,” “Is,” “How,” “Can,” “Which,” and “Others” and then report both per-category accuracies and an overall average score.

Both datasets are based on 3D scenes from the ScanNet dataset~\cite{dai2017scannet}.
In the 3D-VQA task, question answering is typically performed on 3D inputs (which may include images); however, in the proposed method, it is performed on the original RGB-D images from which the 3D data are derived.
Evaluation is performed following the standard protocol defined in each benchmark paper.

\subsection{Implementation Details} 
\label{sec:implementation}

\noindent{\bf Keyframe Extraction} \ The proposed method employs the CLIP-ViT-L/14@336px~\cite{radford2021learning} model as the backbone for vision-language feature extraction. For the keyframe extraction process, we set the parameters to $\alpha = 5.0$ and $\beta = 1.0$, which facilitates a robust selection of representative frames.
We extracted 30 images as keyframes.
We present a sensitivity analysis with respect to the number of images and the parameters $\alpha$ and $\beta$ in Sec.~B of the Supplementary Materials.

\noindent{\bf Prompt generation} \ Because LLMs have difficulty accurately recognizing long-digit numbers, the camera pose is rounded to two decimal places for the position and one decimal place for the rotations.
The keyframe images are resized to a height of 336 px.
In addition, we implement a few-shot prompting strategy as the annotation for benchmark datasets.
The annotation begins with an initial prompt: 
\begin{promptbox} 
Note that the answer for the question is as short as possible such as: 
\end{promptbox} 
\noindent Subsequently, the following template is provided for each question type: 
\begin{promptbox} 
If the question starts with \{qt\} \\
Example answers: \{common\_answer[qt][:20]\}
\end{promptbox} 
\noindent Here, qt represents the question type, whereas common\_answer corresponds to the most frequent answers observed in the training dataset for each question category. 
We supply the top 20 common answers for each question type as the few-shot prompt. 
In ScanQA~\cite{azuma2022scanqa}, the question types include [Where, how many, what color, what shape, what is, and others], and in SQA3D~\cite{ma2022sqa3d}, they are categorized as [What, is, how, can, which, and others]. 
These categorizations mirror the question type classifications defined in each benchmark dataset.
For the complete prompt, please refer to Sec.~C of the Supplementary Materials.

We evaluate the proposed SpatialPrompting framework with two SOTA multimodal LLMs: GPT-4o (gpt-4o-2024-11-20~\cite{openai2024gpt4o}) and Gemini-2.0 (gemini-2.0-flash-exp~\cite{google2024gemini2}).

\subsubsection{Overall Performance}
\label{sec:performance}

\begin{table*}[t]
\begin{center}
\caption{{\bf SpatialQA performances on the ScanQA validation dataset~\cite{azuma2022scanqa} and SQA3D test dataset~\cite{ma2022sqa3d}.} Bold indicates the best performance for a given metric. \textbf{SpatialPrompting} (+ GPT-4o~\cite{openai2024gpt4o}) achieves SOTA in key metrics (EM@1, ROUGE-L, SPICE) on ScanQA, and some question types (What, how, and other) on SQA3D without specialized 3D inputs and 3D-specific fine-tuning.}
\label{tab:quantitative}
\resizebox{\textwidth}{!}{%
\begin{tabular}{l|ccccc|ccccccc}
\hline
                                        & \multicolumn{5}{c|}{ScanQA~\cite{azuma2022scanqa}}                                    & \multicolumn{7}{c}{SQA3D~\cite{ma2022sqa3d}} \\
Method                                  & EM@1        & ROUGE-L     & METEOR      & CIDEr       & SPICE  & What        & Is        & How         & Can       & Which       & Others       & Avg. \\
\hline
ScanQA~\cite{azuma2022scanqa}                & 21.05       & 33.3        & 13.14       & 64.86       & 13.43  & 33.48       & 66.10     & 42.37       & 69.53     & 43.02       & 46.40       & 47.20 \\
3D-VLP~\cite{jin2023context}                 & 21.65       & 34.51       & 13.53       & 66.97       & 14.18  & -           & -         & -           & -         & -           & -           & - \\
3D-LLM~\cite{hong20233d}                     & 20.5        & 35.7        & 14.5        & 69.4        & -      & -           & -         & -           & -         & -           & -           & - \\
LL3DA~\cite{chen2024ll3da}                   & -           & 37.31       & 15.88       & 76.79       & -      & -           & -         & -           & -         & -           & -           & - \\
LEO~\cite{huang2023embodied}                 & -           & 39.3        & 16.2        & 80.2        & -      & -           & -         & -           & -         & -           & -           & 50.0 \\
3D-Vista~\cite{zhu20233d}                    & 27.0        & 38.6        & 15.2        & 76.6        & -      & 34.8        & 63.3      & 45.4        & 69.8      & 47.2        & 48.1        & 48.5 \\
Scene-LLM~\cite{fu2024scene}                 & 27.2        & 40.0        & 16.6        & 80.0        & -      & 40.9        & \bf{69.1} & 45.0        & \bf{70.8} & 47.2        & 52.3        & 54.2 \\
Chat-Scene~\cite{huang2024chat}              & 21.62       & 41.56       & {\bf 18.00} & 87.70       & 20.44  & 45.38       & 67.02     & 52.04       & 69.52     & {\bf 49.85} & 54.95       & 54.57 \\
3DGraphLLM~\cite{zemskova20243dgraphllm}     & -           & -           & -           & 83.1        & -      & -           & -         & -           & -         & -           & -           & \bf{55.2} \\
LSceneLLM~\cite{zhi2024lscenellm}            & -           & 40.82       & 17.95       & {\bf 88.24} & -      & -           & -         & -           & -         & -           & -           & - \\
\hline
\textbf{SpatialPrompting}                    &             &             &             &             &             &             &           &             &        &       &             &    \\
\  \ + GPT-4o~\cite{openai2024gpt4o}         & {\bf 27.34} & {\bf 43.39} & 16.85       & 87.69       & {\bf 20.49} & {\bf 48.65} & 64.26     & {\bf 53.55} & 58.88  & 33.62 & {\bf 55.30} & 52.74 \\
\  \ + Gemini-2.0~\cite{google2024gemini2}   & 26.29       & 41.38       & 15.55       & 80.24       &  18.44      & 42.37       & 60.74     & 46.24       & 57.40  & 40.17 & 48.94       & 48.56 \\
\hline
\end{tabular}
}
\end{center}
\end{table*}

\Cref{tab:quantitative} summarizes the performance of SpatialPrompting on the ScanQA validation dataset~\cite{azuma2022scanqa} and on the SQA3D test dataset~\cite{ma2022sqa3d} alongside several SOTA baselines. Notably, SpatialPrompting (GPT-4o~\cite{openai2024gpt4o}) achieves the highest scores on key metrics such as EM@1, ROUGE-L, and SPICE on ScanQA. Although it does not outperform all competitors on every metric, the proposed method matches or surpasses the SOTA in critical areas---all without requiring specialized 3D inputs or additional fine-tuning.
On SQA3D, SpatialPrompting (GPT-4o~\cite{openai2024gpt4o}) exhibits strong performance, particularly excelling in the “What,” “How,” and “Others” categories. Although its overall average score is marginally lower than those of the SOTA methods, these results demonstrate the robustness of SpatialPrompting in handling diverse question types without 3D-specific fine-tuning. In contrast, the scores in “Is,” “Can,” and “Which" (e.g., “Which direction") categories are inferior to those in other categories. These questions are dependent on the orientation of the user defined by the situation, and we observe that GPT-4o in particular is not good at understanding the orientation of the user based on the situation and answering.

\subsubsection{Ablation Studies} \label{sec:ablation}

\begin{table}[t]
\begin{center}
\caption{{\bf Ablation study on the ScanQA~\cite{azuma2022scanqa} validation dataset and SQA3D~\cite{ma2022sqa3d} test dataset.} We use GPT-4o~\cite{openai2024gpt4o} as the multimodal LLM to evaluate the effectiveness of each component.}
\label{tab:ablation}
\resizebox{\columnwidth}{!}{%
\begin{tabular}{l|ccc|c}
\hline
                                        & \multicolumn{3}{c|}{ScanQA~\cite{azuma2022scanqa}}  & SQA3D~\cite{ma2022sqa3d} \\
Method                                  & EM@1          & ROUGE-L     & CIDEr                 & Avg.                     \\
\hline
{\bf SpatialPrompting (Full)}           & {\bf 27.34}   & {\bf 43.39} & {\bf 87.69}           & 52.74       \\
\ \ w/o KF Extraction                   & 26.40         & 43.00       & 86.78                 & 52.63       \\
\ \ w/o Camera Pose                     & 25.70         & 42.71       & 87.64                 & {\bf 53.05} \\
\ \ w/o Annotation                      & 19.83         & 31.89       & 55.06                 & 47.77       \\
GPT-4o~\cite{openai2024gpt4o} baseline  & 21.43         & 32.18       & 58.83                 & 48.51       \\
\hline
\end{tabular}
}
\end{center}
\end{table}

\begin{figure*}
    \centering
    \includegraphics[width=1.0\linewidth]{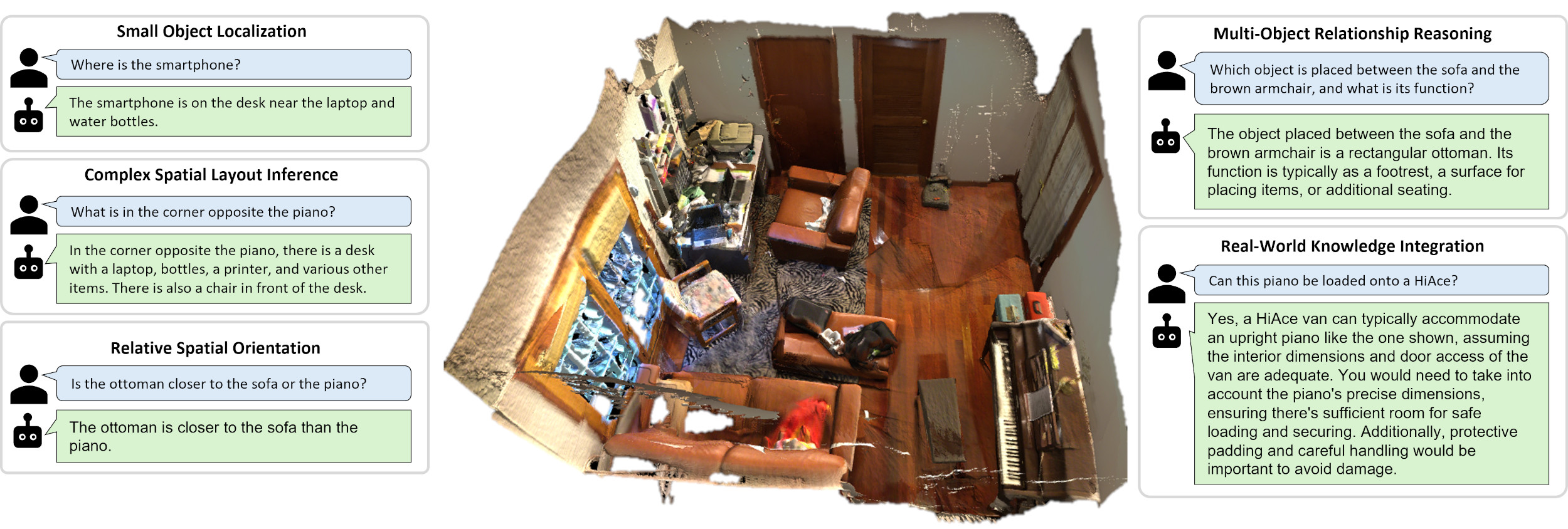}
    \caption{{\bf Qualitative results.} We showcase five categories of spatial reasoning tasks tackled by the proposed SpatialPrompting framework: (1) Small object localization, (2) Complex spatial layout inference, (3) Multi-object relationship reasoning, (4) Relative spatial orientation, and (5) Real-world knowledge integration. Each question--answer pair is generated using minimal keyframes and corresponding camera poses, illustrating how the proposed method handles a range of scenarios---from pinpointing small items to considering practical constraints (e.g., vehicle capacity for transporting a piano).
    }
    \label{fig:qualitative}
\end{figure*}

To evaluate the contribution of each component in the proposed SpatialPrompting framework, we conducted an ablation study on the ScanQA~\cite{azuma2022scanqa} validation dataset and the SQA3D~\cite{ma2022sqa3d} test dataset. 
We specifically examined the effects of three key components: keyframe extraction, camera pose, and the few-shot annotation used to generate our prompt.

For the ablation experiments, we replaced each component with a baseline variant:
\begin{itemize}
\item {\bf Keyframe extraction:} \ Instead of the proposed method for extracting informative keyframes from video, we performed uniform sampling.
\item {\bf Camera Pose:} \ We evaluated the framework without providing camera pose information.
\item {\bf Annotation (Few-shot prompt):} \ We compared our few-shot prompt with a zero-shot variant that instructs the model with the phrase “The answer should be a phrase or a single word.”
\end{itemize}
These variants were thereafter fed into GPT-4o~\cite{openai2024gpt4o}. 

\Cref{tab:ablation} summarizes the results. (Complete results are in Sec.~B.1 of the Supplementary Materials.)
The full SpatialPrompting method achieves the best performance.
Replacing the proposed keyframe extraction method with uniform sampling resulted in a slight decrease across all metrics, indicating that the proposed method better identifies salient frames necessary for spatial reasoning.
Omitting the camera pose information caused a more noticeable drop in performance on ScanQA, indicating that camera pose information is beneficial for spatial context.
In contrast, the score on SQA3D can be marginally improved. This likely stems from a confusion between camera coordinates and user orientation in direction-dependent questions.
In Sec.~B.4 of the Supplementary Materials, we provide an analysis of failure cases.
The removal of few-shot annotations in favor of a zero-shot prompt led to a significant performance degradation, confirming that few-shot prompts are extremely important for guiding LLMs to generate responses tailored to the characteristics of the benchmark datasets.
On the contrary, the naive GPT-4o without SpatialPrompting resulted in considerably lower performance across all metrics, further highlighting the effectiveness of the proposed components.
Overall, the ablation study demonstrates that each component---keyframe extraction, camera pose, and few-shot annotation---contributes to the robust performance of the proposed method.

\subsection{Qualitative Results} 
\label{sec:qualitative}

\begin{figure}
    \centering
    \includegraphics[width=1.0\linewidth]{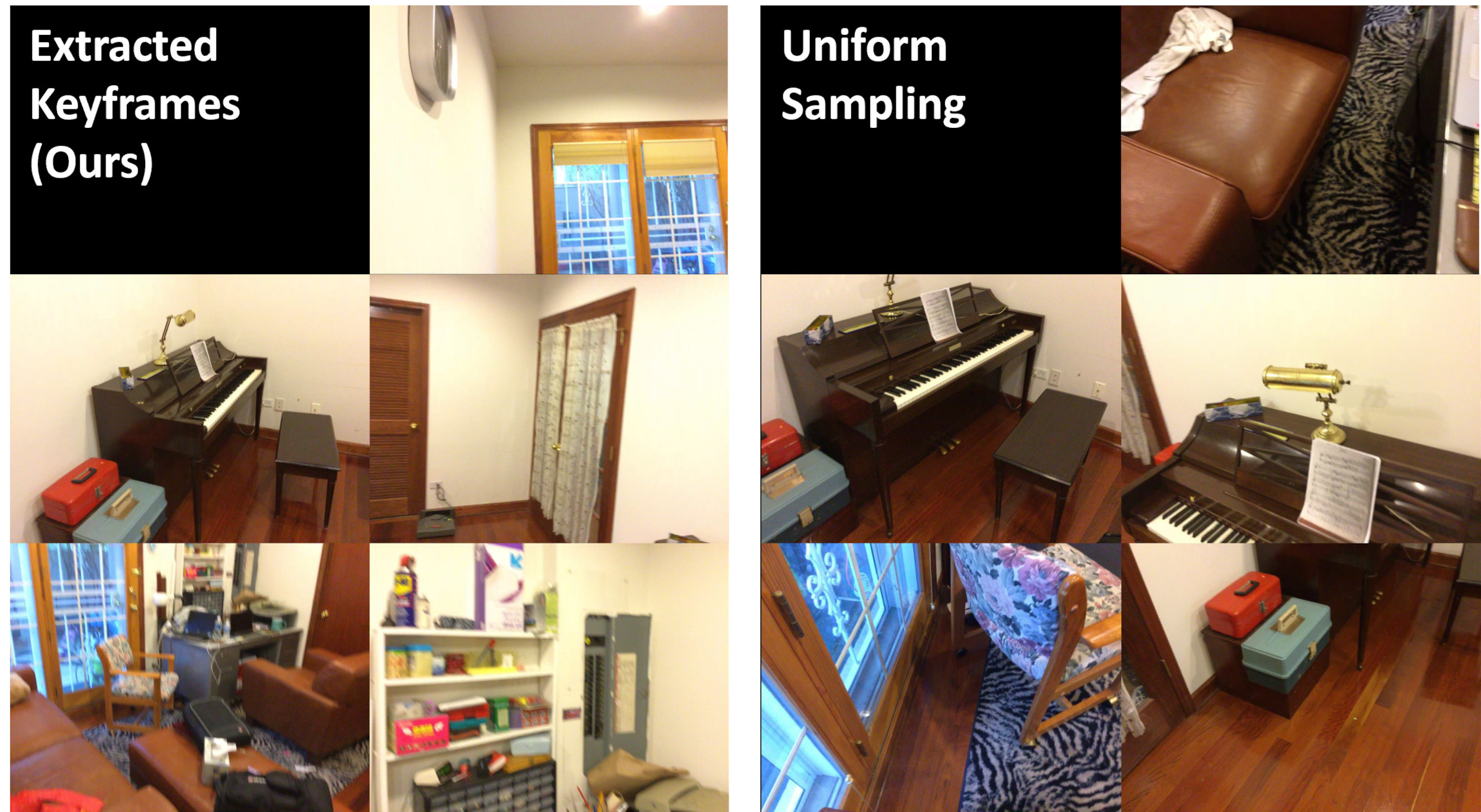}
    \caption{{\bf Effectiveness of Keyframe extraction on SpatialPrompting.} The proposed keyframe extraction process effectively eliminates redundant views, and the results indicate that images with a wider FOV are preferentially selected.}
    \label{fig:keyframe_comparison}
\end{figure}

\begin{figure}
    \centering
    \includegraphics[width=1.0\linewidth]{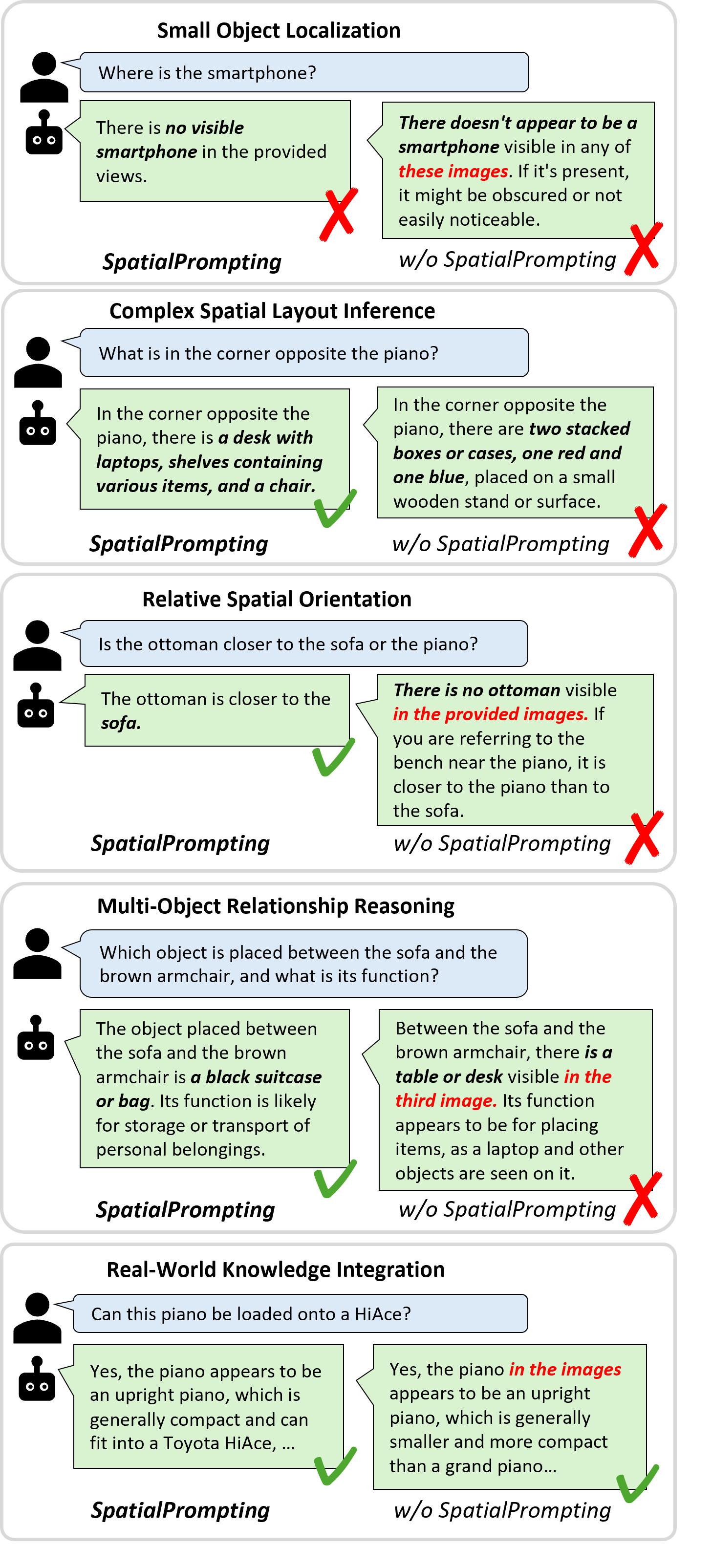}
    \caption{{\bf Effectiveness of SpatialPrompting in limited images.} The use of high-coverage keyframes reduces missing information, whereas camera poses allow the inference of spatial relationships from limited images, enabling the proposed method to achieve complex SpatialQA.}
    \label{fig:qa_examples}
\end{figure}

\Cref{fig:qualitative} demonstrates that SpatialPrompting can effectively handle diverse spatial reasoning tasks by leveraging keyframes and camera pose prompts. Specifically, by inputting raw images, it accurately localizes small objects that are difficult to detect from the 3D model. The introduction of camera pose prompts enables the system to understand complex spatial relationships within a scene. Moreover, by utilizing a pre-trained LLM enriched with extensive world knowledge, the proposed method naturally combines scene-specific cues with general knowledge to provide coherent answers, all without any special fine-tuning.

Next, to demonstrate the effectiveness of the proposed keyframe extraction, a visual comparison with uniform sampling is presented in \cref{fig:keyframe_comparison}.
The proposed method prioritizes frames with wider FOVs, higher sharpness, and diverse viewpoints, whereas the baseline simply samples images at regular intervals. Consequently, the keyframes selected by the proposed algorithm capture a more comprehensive view of the scene, including critical objects, spatial layouts, and angles that are often overlooked by uniform sampling.

\Cref{fig:qa_examples} demonstrates a case in which the proposed SpatialPrompting approach leverages the five images described above to illustrate the benefits of the proposed method for QA tasks. 
When asked, “Where is the smartphone?,” both methods fail to detect the smartphone owing to the limited coverage in the selected views. 
For queries such as “What is in the corner opposite the piano?,” SpatialPrompting correctly identifies the appropriate corner by utilizing the camera pose, whereas the baseline method misinterprets the corner, resulting in an incorrect answer. 
In the subsequent two questions, the narrow FOV resulting from uniform sampling causes failures in answering, whereas the wide FOV achieved through keyframe selection in SpatialPrompting addresses this issue. 
For a question requiring both spatial understanding and real-world knowledge---namely, “Can this piano be loaded onto a HiAce?”---both methods perform successfully, as the existing multimodal LLM, such as GPT-4o, already integrates real-world knowledge. 
Focusing on the content of the responses, we observed that the naive version of GPT-4o frequently employed expressions that referred to the provided images. This indicates that when the proposed annotations are not used, the LLMs generate responses that directly reference images unknown to the user, thereby compromising the user experience.
We provide additional qualitative evaluation in Sec.~A of the Supplementary Materials.
Therefore, SpatialPrompting facilitates a deeper understanding of the spatial context even with few images.

\section{Discussion} 
\label{sec:discussion}
Our experimental results highlight several strengths of SpatialPrompting:
\begin{itemize} 
\item {\bf Versatility:} The method achieves consistent performance across diverse SpatialQA tasks, comparable to results from more specialized approaches. 
\item {\bf Efficiency:} By eliminating the need for 3D-specific fine-tuning, SpatialPrompting simplifies system architecture and reduces both training costs and data requirements. 
\item {\bf Spatial reasoning:} The approach effectively leverages the latent spatial understanding inherent in SOTA multimodal LLMs (e.g., GPT-4o). 
\end{itemize}

However, our analysis of SQA3D~\cite{ma2022sqa3d} in \cref{sec:performance} reveals limitations in addressing questions involving directional cues.
Our findings, which align with observations by Deguchi et al.~\cite{deguchi2024language}, indicate that current LLMs may not sufficiently capture the situated context (e.g., user posture).
Addressing this limitation by incorporating additional spatial descriptors or integrating dedicated modules for interpreting user context represents an important avenue for future studies.
Another limitation is that SpatialPrompting does not explicitly estimate metric details, such as the precise positions of objects, which may necessitate dedicated estimation modules.

\section{Conclusion} 
\label{sec:conclusion}
This study introduces SpatialPrompting, a novel framework for zero-shot spatial reasoning. The proposed approach replaces expensive 3D-specific fine-tuning with a keyframe-driven prompt generation strategy, leveraging representative keyframes and camera pose data. Evaluations on challenging benchmarks, including ScanQA~\cite{azuma2022scanqa} and SQA3D~\cite{ma2022sqa3d}, demonstrate that SpatialPrompting delivers robust performance while maintaining a simple and scalable architecture. 
Although challenges remain in capturing fine-grained spatial nuances, the proposed method offers significant advantages in terms of efficiency and scalability. 
We believe that the flexibility of SpatialPrompting opens up a rich landscape of research opportunities, ranging from enhanced spatial context interpretation to potential applications in diverse domains. 
This study is expected to stimulate further exploration and inspire the community to pursue novel approaches in robust spatial reasoning.

{
    \small
    \bibliographystyle{ieeenat_fullname}
    \bibliography{main}
}

\clearpage

\clearpage

\appendix
\onecolumn

\noindent
{\LARGE \bf Supplementary Materials}

\section{Additional Qualitative Results} \label{appendix:additional_qualitative}
\begin{figure}[h]
    \centering
    \includegraphics[width=1.0\linewidth]{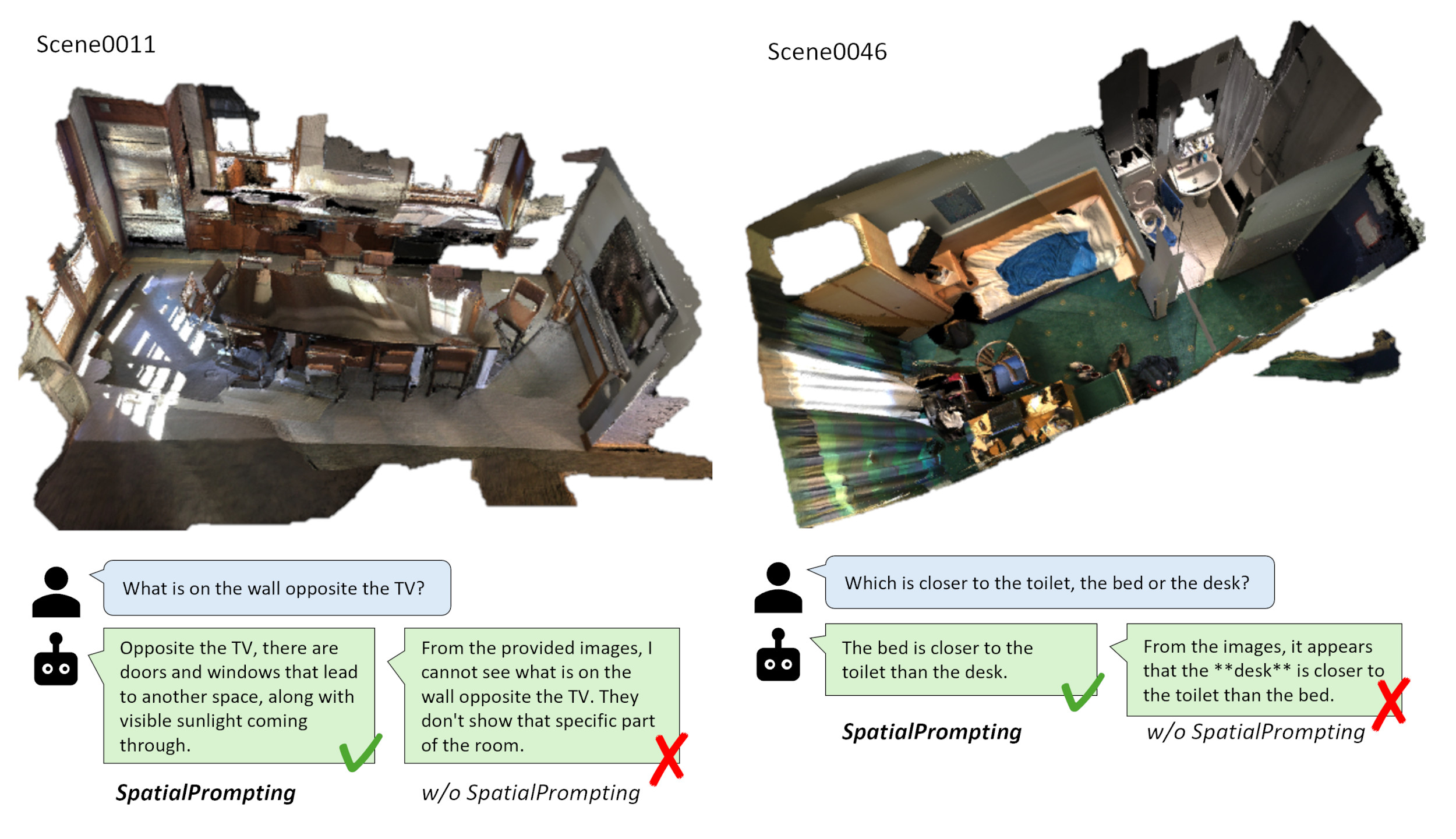}
    \caption{{\bf Additional Qualitative Result on ScanNet~\cite{dai2017scannet} (1).} We use GPT-4o~\cite{openai2024gpt4o} as the multimodal LLM and inputs five images for SpatialQA.}
    \label{fig:additional_qualitative}
\end{figure}

\begin{figure}[h]
    \centering
    \includegraphics[width=1.0\linewidth]{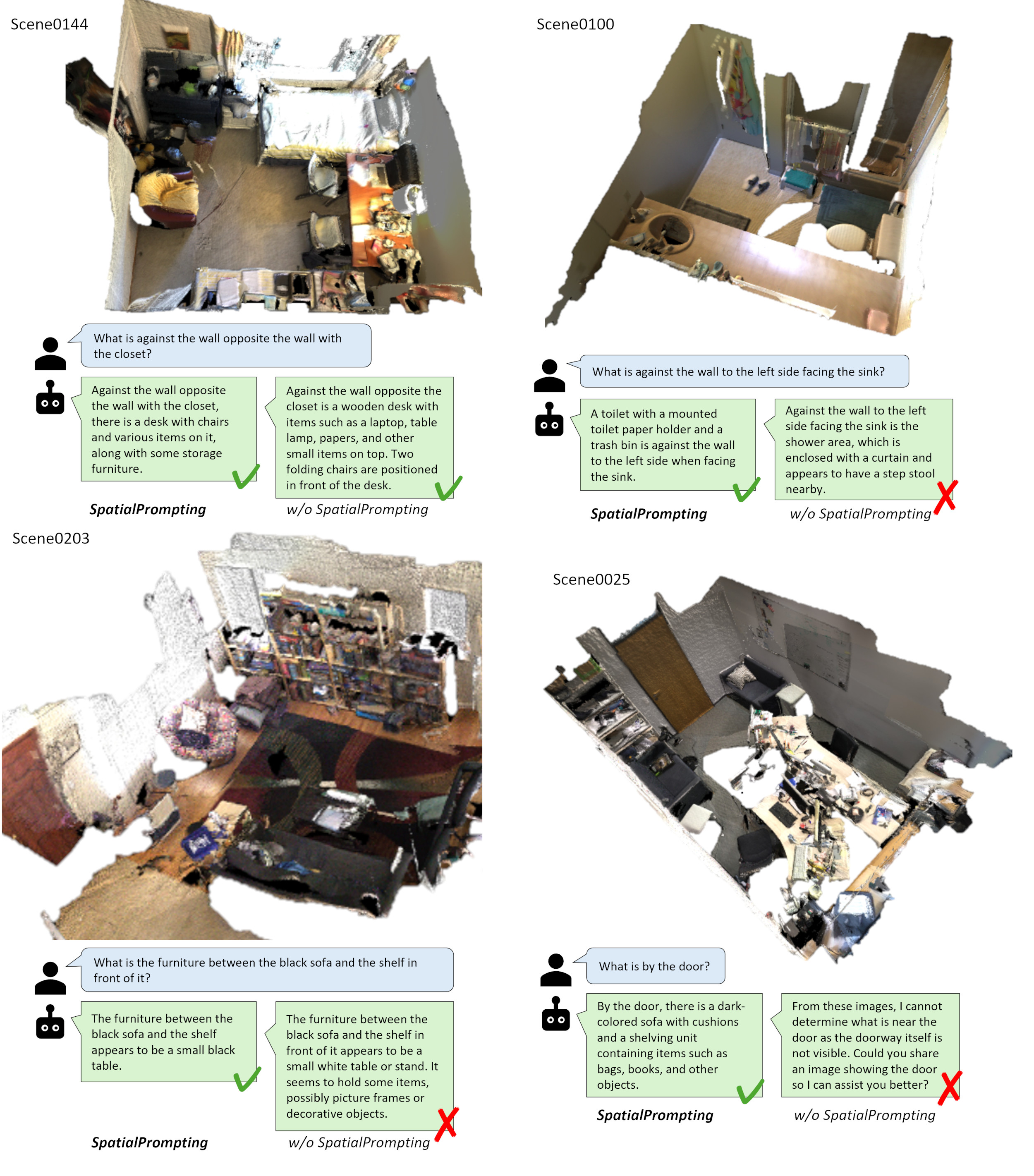}
    \caption{{\bf Additional Qualitative Result on ScanNet~\cite{dai2017scannet} (2).} We use GPT-4o~\cite{openai2024gpt4o} as the multimodal LLM and inputs five images for SpatialQA.}
    \label{fig:additional_qualitative2}
\end{figure}

In this section, we present further qualitative results of our SpatialPrompting method.
Figures~\ref{fig:additional_qualitative} and \ref{fig:additional_qualitative2} provide additional examples to show that our method effectively handles complex spatial reasoning tasks in various indoor scenes. Specifically, this example shows that our method can infer complex spatial layouts that cannot be obtained from a single image. Selected keyframes, along with corresponding camera pose data, capture important spatial cues that uniform sampling methods often miss. These qualitative results highlight the spatial reasoning ability of our method in real-world scenarios.
Conversely, some results indicate that GPT-4o~\cite{openai2024gpt4o} can infer spatial relationships solely from only images, which are not deducible from a single image, suggesting that LLMs may possess intrinsic spatial inference capabilities.

\clearpage
\section{Extended Ablation Studies} \label{appendix:ex_ablation}

\subsection{Complete Ablation Study Results on Benchmark Datasets} \label{appendix:complete_ablation}

Tables~\ref{tab:ablation_scanqa} and \ref{tab:ablation_sqa3d} present the complete ablation study results on the benchmark datasets, ScanQA~\cite{azuma2022scanqa} and SQA3D~\cite{ma2022sqa3d}. The results clearly indicate that the removal of the proposed component leads to a degradation in performance. However, as discussed in Sec.\ref{sec:ablation}, we observed that for certain questions in SQA3D\cite{ma2022sqa3d}—specifically those relying on the user's orientation (e.g., question types “Is,” “Can,” “Which,” and “Others”)—the model sometimes confuses the camera coordinates with the user’s orientation.

\begin{table}[h]
\begin{center}
\caption{{\bf Ablation Study on the ScanQA \cite{azuma2022scanqa} Validation Dataset.} We use GPT-4o~\cite{openai2024gpt4o} as the multimodal LLM.}
\label{tab:ablation_scanqa}
\resizebox{\textwidth}{!}{%
\begin{tabular}{lccccccccccc}
\hline
Method                                  & EM@1          & B-1           & B-2         & B-3         & B-4         & ROUGE-L       & METEOR      & CIDEr       & SPICE       \\
\hline
{\bf SpatialPrompting (Full)}           & {\bf 27.34}   & 38.49         & 25.09       & 16.61       & {\bf 10.85} & {\bf 43.39}   & {\bf 16.85} & {\bf 87.69} & 20.49       \\
\ \ w/o KF Extraction                   & 26.40         & {\bf 38.65}   & 25.17       & 16.24       & 9.84        & 43.00         & 16.76       & 86.78       & 20.49       \\
\ \ w/o Camera Pose                     & 25.70         & 38.33         & {\bf 25.43} & {\bf 17.18} & 9.71        & 42.71         & 16.79       & 87.64       & {\bf 21.26} \\
\ \ w/o Annotation                      & 19.83         & 20.56         & 7.41        & 3.25        & 1.51        & 31.89         & 12.16       & 55.06       & 14.17       \\
GPT-4o~\cite{openai2024gpt4o} baseline  & 21.43         & 26.37         & 9.65        & 4.07        & 1.71        & 32.18         & 11.80       & 58.83       & 13.55       \\
\hline
\end{tabular}
}
\end{center}
\end{table}

\begin{table}[h]
\begin{center}
\caption{{\bf Ablation Study on the SQA3D\cite{ma2022sqa3d} Test Dataset.} We use GPT-4o~\cite{openai2024gpt4o} as the multimodal LLM.}
\label{tab:ablation_sqa3d}
\begin{tabular}{lccccccc}
\hline
Method                                  & \multicolumn{6}{c}{\it Test set}                                                  & Avg. \\
\cline{2-7}
                                        & What        & Is        & How         & Can       & Which       & Others           & \\
\hline
{\bf SpatialPrompting (Full)}           & 48.65       & 64.26       & {\bf 53.55} & 58.88       & 33.62       & 55.30       & 52.74 \\
\ \ w/o KF Extraction                   & {\bf 49.61} & 64.88       & 51.83       & 57.10       & 32.19       & 55.30       & 52.63 \\  
\ \ w/o Camera Pose                     & 48.47       & {\bf 65.64} & 52.04       & {\bf 60.06} & 35.04       & {\bf 55.65} & {\bf 53.05} \\
\ \ w/o Annotation                      & 39.23       & 62.12       & 48.82       & 52.37       & 40.46       & 49.47       & 47.77 \\
GPT-4o~\cite{openai2024gpt4o} baseline  & 40.10       & 63.96       & 45.16       & 52.67       & {\bf 44.44} & 50.53       & 48.51 \\
\hline
\end{tabular}
\end{center}
\end{table}

\subsection{Sensitivity Analysis of the Number of Input Images} \label{appendix:sensitivity_num_images}
\Cref{tab:sensitivity_num_images} presents an analysis of how varying the number of input images affects performance on ScanQA validation dataset~\cite{azuma2022scanqa}. 
The results indicate that using the maximum number of images (30) provides the best balance between comprehensive scene coverage and overall model performance. 
As the number of images is reduced (to 15 or 5), key metrics such as EM@1 and ROUGE-L show a slight decline, suggesting that richer visual inputs yield more robust spatial inference. 
However, since incorporating a larger number of images incurs higher API costs and latency, the optimal number of images should be chosen based on system requirements. Additionally, experiments confirm that the SpatialPrompting enhances performance across all tested configurations.

\begin{table}[h]
\begin{center}
\caption{{\bf Sensitivity Analysis of Number of Images on the ScanQA \cite{azuma2022scanqa} Validation Dataset.} We use GPT-4o~\cite{openai2024gpt4o} as the multimodal LLM.}
\resizebox{\textwidth}{!}{%
\label{tab:sensitivity_num_images}
\begin{tabular}{lccccccccccc}
\hline
Method                                 & \# Images   & EM@1          & B-1           & B-2         & B-3         & B-4         & ROUGE-L       & METEOR      & CIDEr       & SPICE \\
\hline
{\bf SpatialPrompting}                 &  30         & {\bf 27.34}   & {\bf 38.49}   & 25.09       & 16.61       & {\bf 10.85} & {\bf 43.39}   & {\bf 16.85} & {\bf 87.69} & {\bf 20.49} \\
                                       &  15         & 26.63         & 38.11         & {\bf 25.14} & {\bf 17.46} & 10.59       & 42.64         & 16.63       & 85.69       & 20.37 \\
                                       & 5           & 23.77         & 35.16         & 22.91       & 15.59       & 9.67        & 39,07         & 15.24       & 77.86       & 18.45 \\
\hline
GPT-4o~\cite{openai2024gpt4o} baseline &  30         & 21.43         & 26.37         & 9.65        & 4.07        & 1.71        & 32.18         & 11.80       & 58.83       & 13.55 \\
                                       & 15          & 21.43         & 26.06         & 9.72        & 4.28        & 2.03        & 31.91         & 11.67       & 58.34       & 13.69 \\
                                       & 5           & 18.59         & 18.52         & 6.40        & 2.66        & 1.08        & 28.18         & 10.35       & 50.20       & 12.28 \\
\hline
\end{tabular}
}
\end{center}
\end{table}

\clearpage
\subsection{Sensitivity Analysis of Parameters} \label{appendix:sensitivity_parameters}
\begin{figure}[h]
    \centering
    \includegraphics[width=1.0\linewidth]{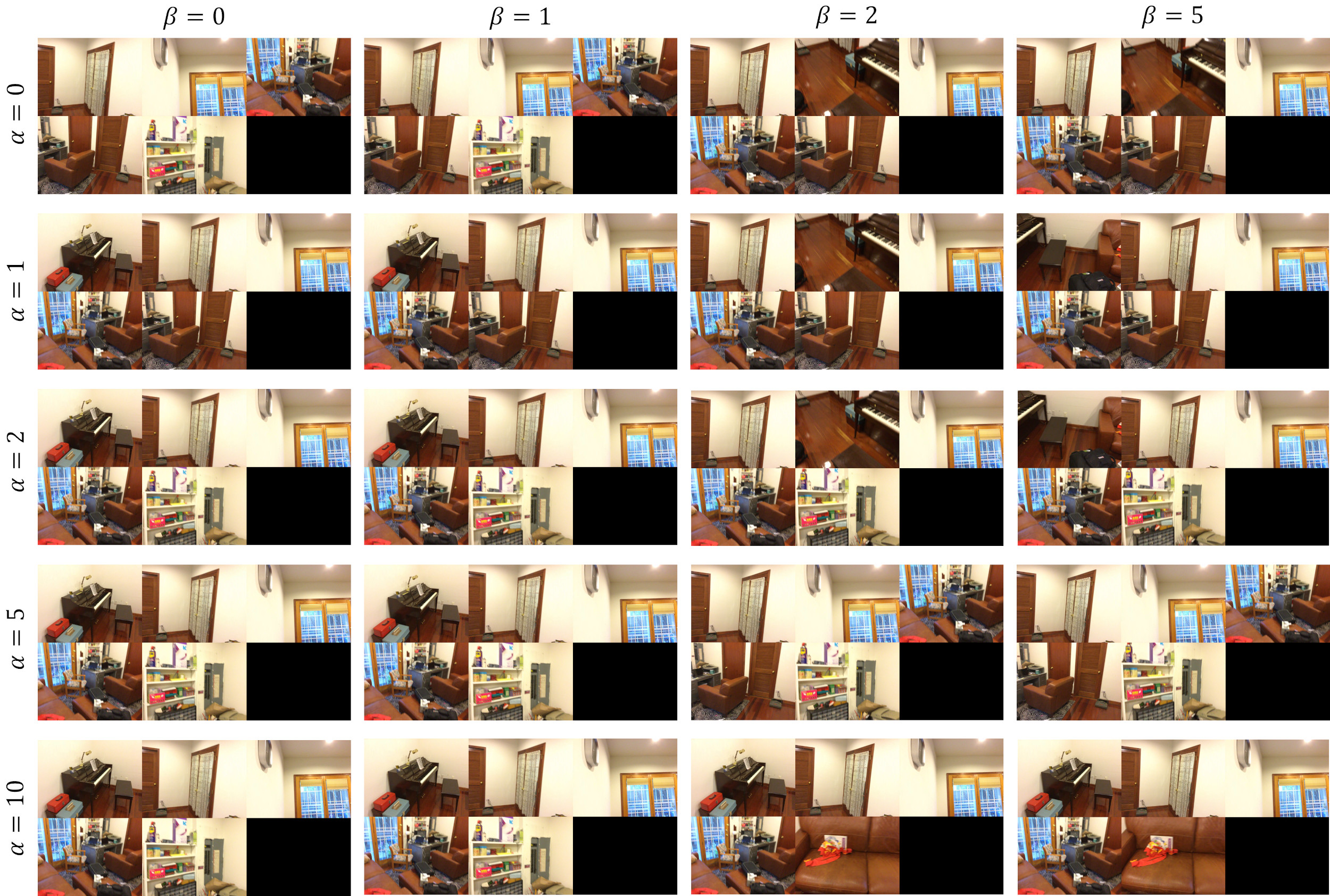}
    \caption{{\bf Sensitivity Analysis of Parameters $\alpha$ and $\beta$}}
    \label{fig:sensitivity_parameters}
\end{figure}

In \cref{fig:sensitivity_parameters}, we present five extracted keyframes obtained by varying the parameters $\alpha$ and $\beta$. A small value for $\alpha$, which considers primarily semantic distance, can lead to the overlooking of distinctive features—such as the piano—or result in the selection of similar images with a broad field of view. Conversely, setting $\alpha$ too high may compromise spatial coverage. Similarly, when $\beta$ is omitted, there is an increased likelihood of selecting blurred images; however, an excessively high $\beta$ can cause the algorithm to favor images with an overly narrow spatial perspective. Based on these considerations and through iterative experimentation, we determined that setting $\alpha = 5.0$ and $\beta = 1.0$ produced the most balanced results.

\clearpage
\subsection{Failure Analysis} \label{appendix:failuer_analysis}
In this section, we describe representative failure cases of SpatialPrompting on the SQA3D dataset~\cite{ma2022sqa3d}.

\begin{figure}[h]
    \centering
    \includegraphics[width=1.0\linewidth]{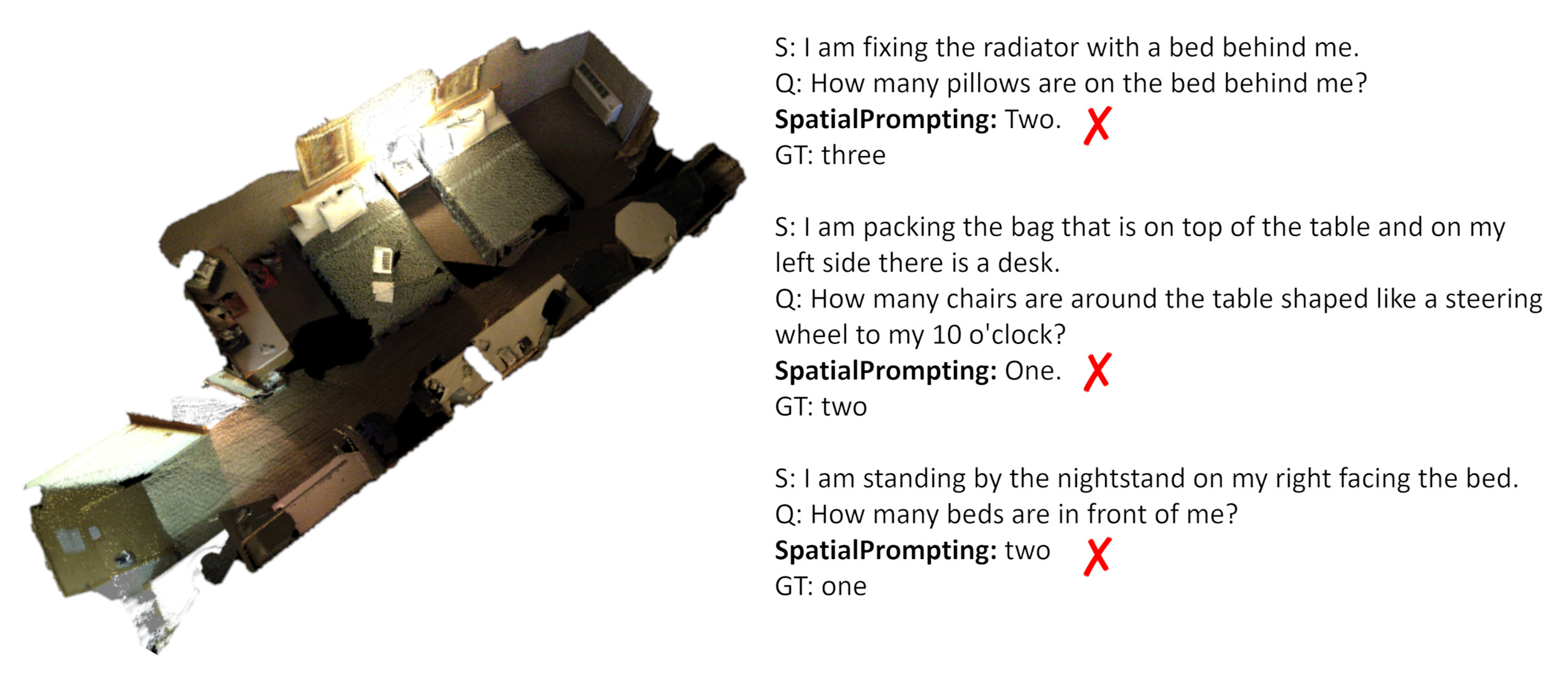}
    \caption{{\bf Examples of the Miscount of Objects on SQA3D~\cite{ma2022sqa3d}.} S represents the situation, Q represents the question, SpatialPrompting denotes the predicted answer generated by our SpatialPrompting method using GPT-4o~\cite{openai2024gpt4o}, and GT denotes the ground-truth answer.}
    \label{fig:failure_counting}
\end{figure}

\noindent{\bf Miscount of Objects} \ One common failure case is miscounting of objects, as illustrated in \cref{fig:failure_counting}. Multimodal LLMs have inherent limitations in their ability to count objects during image understanding, and since SpatialPrompting delegates object counting to the multimodal LLM, it inherits these same limitations. If more precise object counting information becomes available, incorporating it may lead to improved performance.

\begin{figure}[h]
    \centering
    \includegraphics[width=1.0\linewidth]{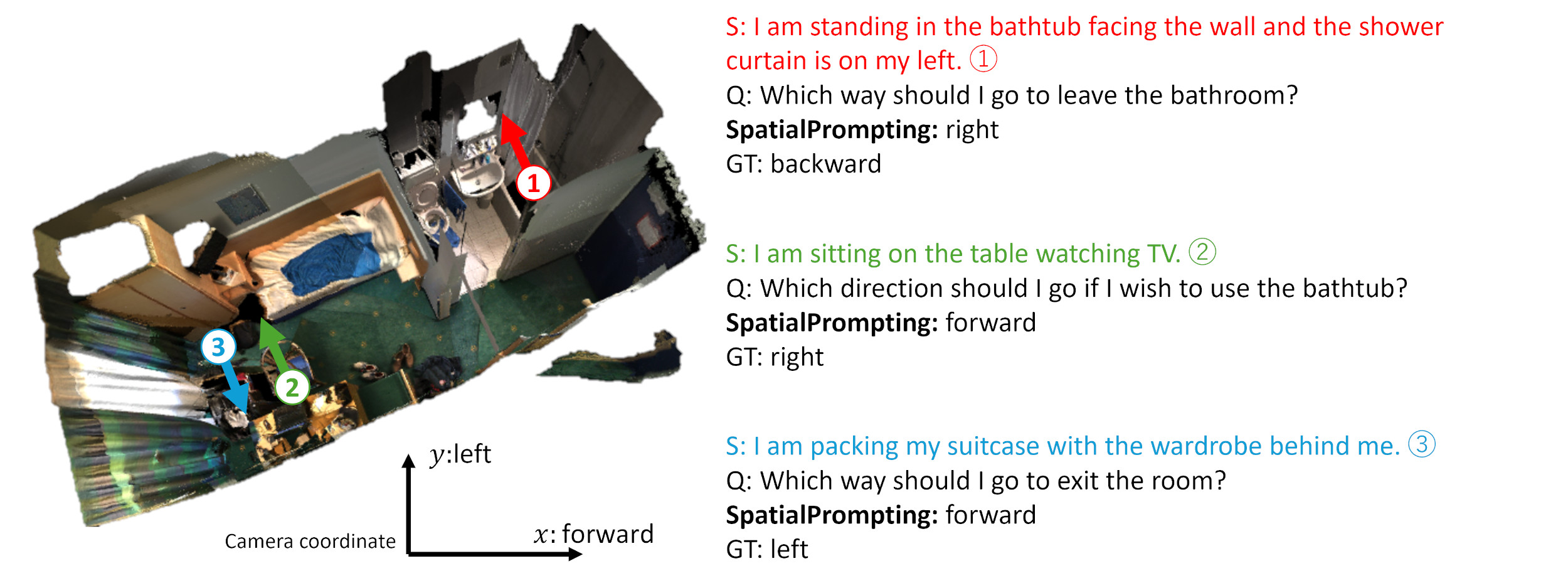}
    \caption{{\bf Examples of the Misdirection on SQA3D~\cite{ma2022sqa3d}.} S represents the situation, Q represents the question, SpatialPrompting denotes the predicted answer generated by our SpatialPrompting method using GPT-4o~\cite{openai2024gpt4o}, and GT denotes the ground-truth answer.}
    \label{fig:misdirection}
\end{figure}

\noindent{\bf Misdirection} \ It has been suggested that directional errors are prevalent in SQA3D~\cite{ma2022sqa3d}, as exemplified in \cref{fig:misdirection}. Although SpatialPrompting is provided with camera poses, several questions in SQA3D require answers based on the user's orientation. 
In these tasks, we observed several scenes where the system appears to conflate the user's orientation with the given camera coordinates, leading to responses derived from the camera’s orientation instead of the user’s actual orientation.

\begin{figure}[h]
    \centering
    \includegraphics[width=1.0\linewidth]{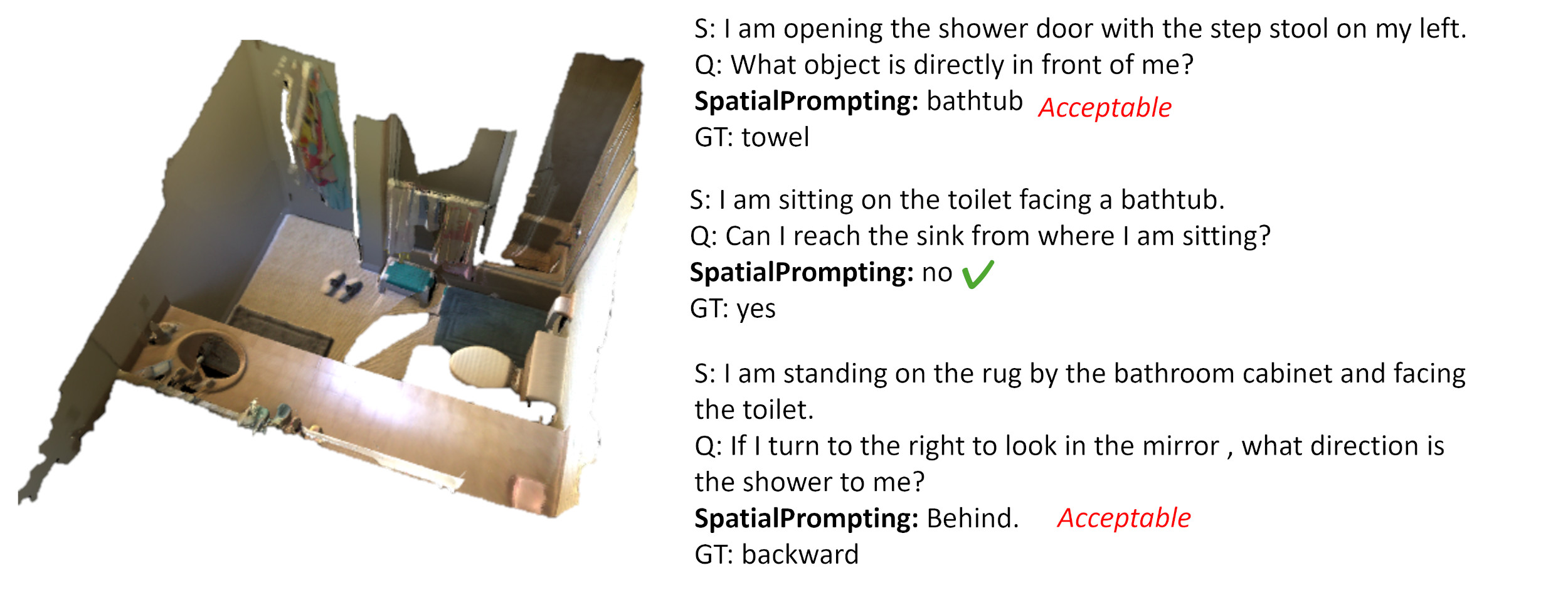}
    \caption{{\bf Examples of Acceptable Variants on SQA3D~\cite{ma2022sqa3d}.} S represents the situation, Q represents the question, SpatialPrompting denotes the predicted answer generated by our SpatialPrompting method using GPT-4o~\cite{openai2024gpt4o}, and GT denotes the ground-truth answer.}
    \label{fig:acceptable_variants}
\end{figure}

\clearpage
\noindent{\bf Acceptable Variants} \ Some of the failure examples are actually acceptable variants, as illustrated in \cref{fig:acceptable_variants}. For example, the top case is deemed acceptable due to the ambiguity of expressions about the situation. In the middle case, even though the ground-truth is "yes," the answer indicating that the hand cannot reach the sink from the toilet seems more appropriate. Lastly, the third example results from differences in expression; because SQA3D evaluates only by exact match, such cases are also counted as failures.

\clearpage
\section{Additional Implementation Details} \label{appendix:additional_implementation_details}

\subsection{Complete Few-shot Prompt for Benchmark Datasets}
We implement a few-shot prompting strategy as the annotation for benchmark datasets, ScanQA~\cite{azuma2022scanqa} and SQA3D~\cite{ma2022sqa3d}.
The annotation is consists of the top 20 frequent answers for each question type in the training datasets.

\noindent{\bf ScanQA~\cite{azuma2022scanqa}} \ The complete few-shot prompt for ScanQA~\cite{azuma2022scanqa} is consists of the answers for the question types include [Where, How many, What color/What is the color, What shape/What type/What kind, What is, others] as follows:

\begin{promptbox}
Note that the answer for the question is as short as possible such as: \\
\\
If question is start with Where \\
Example of answers: against wall, on desk, yes, under desk, on wall, center of room, under window, in corner, in front of window, middle of room, in middle of room, left, under table, to left, to right, at table, above sink, above toilet, left of toilet, right of toilet,  \\
\\
If question is start with How many \\
Example of answers: 2, 3, 4, 1, 5, 6, 4 chairs, 2 chairs, 8, 3 chairs, 4 legs, 5 chairs, 1 chair, 2 monitors, yes, 2 pillows, 2 stools, 2 printers, 2 seats, 2 sinks,  \\
\\
If question is start with What color, What is the color \\
Example of answers: white, brown, black, blue, grey, red, tan, light brown, gray, beige, green, dark brown, yes, silver, yellow, black chair, orange, white door, black and white, brown chair, \\
\\
If question is start with What shape, What type, What kind \\
Example of answers: rectangular, rectangle, square, rectangular shape, round, wooden, yes, oval, white, whiteboard, black, office chair, rectangular door, black office chair, circular, coffee table, cylindrical, glass, rectangular window, white board,  \\
\\ 
If question is start with What is \\
Example of answers: chair, table, trash can, window, desk, door, couch, picture, whiteboard, cabinet, black chair, shelf, yes, radiator, bed, coffee table, lamp, sink, chairs, tv, \\
\\
If question is start with others \\
Example of answers: right, left, table, right side, trash can, yes, chair, window, left side, desk, wall, couch, door, cabinet, center, radiator, to right, bookshelf, chairs, shelf, \\
\\
\end{promptbox}

\clearpage
\noindent{\bf SQA3D~\cite{ma2022sqa3d}} \ The complete few-shot prompt for SQA3D~\cite{ma2022sqa3d} is consists of the answers for the question types include [What, Is, How, Can, Which, Others] as follows:

\begin{promptbox}
Note that the answer for the question based on the situation is as short as possible such as: \\
\\
If question is start with What \\
Example of answers: brown, white, table, black, window, chair, door, couch, picture, bed, cabinet, backpack, shelf, desk, whiteboard, trash can, blue, red, radiator, sink,  \\
\\
If question is start with Is \\
Example of answers: yes, no, even, odd, rectangular, closed, open, left, right, messy, off, behind, on, square, round, down, tidy, large, up, neat, \\
\\
If question is start with How \\
Example of answers: one, two, three, four, zero, five, backward, six, forward, right, left, eight, seven, in rows, 180 degrees, nine, turn around, wall mounted, ten, stacked, \\
\\
If question is start with Can \\
Example of answers: yes, no, TV, backward, behind, black, copier, f, hear, jump, left, no ,you will be blocked by desk, one, open it first, right, two, walk forward, yes, but i need to turn my head slightly to right, yes, turn right, \\
\\
If question is start with Which \\
Example of answers: right, left, backward, forward, behind, couch, table, bed, TV, chair, front, window, coffee table, sink, backpack, bathtub, desk, door, trash can, recycling bin, \\
\\
If question is start with Others \\
Example of answers: yes, no, right, left, backward, forward, even, behind, true, closed, odd, window, two, chair, sink, table, open, bed, whiteboard, false, \\
\\  
\end{promptbox}

\end{document}